\documentclass[sigconf,letterpaper,9pt]{acmart}
\acmConference[FAT*'19]{ACM Conference on Fairness, Accountability, and
Transparency}{January 2019}{Atlanta, Georgia USA}
\acmYear{2019}
\copyrightyear{2019}

\usepackage{booktabs} 

\usepackage{amssymb}
\usepackage{amsthm}

\usepackage{tikz}
\usetikzlibrary{
  shapes,calc,arrows,fit,positioning,decorations.pathmorphing,snakes,intersections,shapes.geometric,trees,cd
}

\begin{document}

\ccsdesc[500]{Social and professional topics~Race and ethnicity}
\ccsdesc[500]{Applied computing~Sociology}
\ccsdesc[300]{Social and professional topics~Systems analysis and design}
\ccsdesc[300]{Computing methodologies~Dimensionality reduction and manifold learning}

\keywords{fairness, machine learning, racial classification, segregation}

\copyrightyear{2019} 
\acmYear{2019} 
\setcopyright{acmcopyright}
\acmConference[FAT* '19]{FAT* '19: Conference on Fairness, Accountability, and Transparency}{January 29--31, 2019}{Atlanta, GA, USA}
\acmBooktitle{FAT* '19: Conference on Fairness, Accountability, and Transparency (FAT* '19), January 29--31, 2019, Atlanta, GA, USA}
\acmPrice{15.00}
\acmDOI{10.1145/3287560.3287575}
\acmISBN{978-1-4503-6125-5/19/01}

\title{Racial categories in machine learning}
\titlenote{Produces the permission block, and
  copyright information}

\author{Sebastian Benthall}
\affiliation{%
  \institution{New York University}
}

\author{Bruce D. Haynes}
\affiliation{%
  \institution{University of California, Davis}
}


\begin{abstract}
  Controversies around race and machine learning have sparked
  debate among computer scientists over how to design
  machine learning systems that guarantee fairness.
  These debates rarely engage with how racial identity is
  embedded in our social experience, making for sociological
  and psychological complexity. This complexity challenges the
  paradigm of considering fairness to be a formal property of
  supervised learning with respect to protected personal attributes.
  Racial identity is not simply a personal subjective quality.
  For people labeled ``Black'' it is an ascribed political category
  that has consequences for social differentiation embedded in
  systemic patterns of social inequality achieved through both
  social and spatial segregation. In the United States,
  racial classification
  can best be understood as a system of inherently unequal status
  categories that places whites as the most privileged category
  while signifying the Negro/black category as stigmatized.
  Social stigma is reinforced through the unequal distribution of
  societal rewards and goods along racial lines that is reinforced
  by state, corporate, and civic institutions and practices.
  This creates a dilemma for society and
  designers: be blind to racial group disparities and thereby
  reify racialized social inequality by no longer measuring systemic
  inequality, or be conscious of racial categories in a way that
  itself reifies race. We propose a third option. By preceding group
  fairness interventions with unsupervised learning to dynamically
  detect patterns of segregation, machine learning systems
  can mitigate the root cause of social disparities,
  social segregation and stratification, without further anchoring
  status categories of disadvantage. 
\end{abstract}

\maketitle

A growing community of researchers and practitioners now
studies fairness in applications of machine learning in such sensitive
areas as credit reporting, employment, education, criminal justice,
and advertising.
This scholarship has been motivated by pragmatic concerns about
machine-learning-produced group biases and compliance
with nondiscrimination law, as well as a general concern about social
fairness. While many of the controversies that have inspired this
research have been about discriminatory impact
on particular groups, such as  Blacks or women, computer scientists
have tended to treat group fairness abstractly, in terms of generic
protected classes, rather than in terms of specific status groups.
This leads analysts to treat ranked racial and gender status
categories simply as nominal categories of personal identity
(a characteristic of the individual) in computational analysis,
rather than understanding that male/female or Negro (black)/white
are each systems of hierarchical social statuses. 

The typical literature in this field addresses problems in a
supervised machine learning paradigm, wherein a predictor is
trained on a set of personal data $X$. One or more features or
columns of the personal data, $A$, are protected demographic
categories.
Each person is labeled with the desired outcome value $Y$,
and a classifier or predictor $\hat{Y}$ is trained on the
labeled data set.
The data is assumed to be accurate. Fairness is then defined as a
formal property of the predictor or prediction algorithm, defined in terms of the training data. Several different formal definitions of fairness have been proposed, and their relationships with each other are well studied.
Some proposed definitions of fairness are \cite{kusner2017counterfactual}: 

\begin{definition}[Fairness through unawareness (FTU)]
  An algorithm is fair so long as protected attributes $A$
  are not explicitly used in the decision-making.
\end{definition}

\begin{definition}[Demographic parity (DP)]
  A predictor $\hat{Y}$ satisfies demographic parity
  if $P(\hat{Y} \vert A = 0) = P(\hat{Y} \vert A = 1)$
\end{definition}

\begin{definition}[Equality of Opportunity (EO)]
  A predictor $\hat{Y}$ satisfies equality of opportunity if
  if $P(\hat{Y} = 1\vert A = 0, Y = 1) = P(\hat{Y} = 1\vert A = 1, Y = 1)$
\end{definition}

Comparatively little attention is given to how the protected class
labels, A, are assigned, why they are
being protected and by whom, and what that means for the normative
presumptions typical of ``fair'' machine learning design.
This paper addresses these questions with focus on the particular but
also paradigmatic
case in which the protected class is a specific racial
category--African-American (Black). We argue, using this case,
that rather than being an
abstract, nominal category, race classification is
embedded in state institutions,
and reinforced in civil society in ways that are relevant to the
design of machine learning systems.
Research demonstrates that race
categories have been socially constructed as unequal categories in
numerous Latin American nations and in the United States.
Race provokes discussions of fairness because racial classification
signifies social, economic, and political inequities  anchored in
state, and civic institutional practices.

Racial categories are
also unstable social constructions as a brief history of race since
late nineteenth-century America will reveal. We will show how race
categories have been subject to constant political contestation
in meaning and  as a consequence racial identity itself is in
fact not stable. Race is ascribed onto individual bodies at
different times and places in society based on many variables
including specific ocular-corporeal characteristics, social class,
perceived ancestral origins,
and state policy \cite{loveman2014national,roth2012race}.   

As a consequence of the social and historical facts about
racial classification, many machine learning applications
that perform statistical profiling, and especially
those that use racial statistics, are both technically
and politically problematic.
Because race is not an inherent property of
a person but a `social fact' about their political place
and social location in society, racial statistics do not
reflect a stable `ground truth'. Moreover, racial statistics
by their very nature mark a status inequality, a way of
sorting people's life chances, and so are by necessity
correlated with social outcomes. There is nothing fair about
racial categories. Scholars of ``fairness in machine learning''
using racial categories should be reflexive about this paradox.

The social facts about race present a dilemma for system
designers. Systems that learn from broad population data
sets without considering racial categories will reflect
the systemic racial inequality of society.
Through their effects on resource allocation, these systems
will reify these categories by disparately impacting racially
identified groups. On the other hand, systems that explicitly
take racial classification into account must rely on individual
identification and/or social ascription of racial categories
that are by definition unequal. Even when these
classifications are used in a ``fair'' way, they reify
the categories themselves.

We present a third option as a potential solution to
this dilemma. The history of racial formation shows that
the social fact of racial categorization is reinforced
through policies and practices of segregation and
stratification in housing, education, employment, and civic life.
Racial categories are ascribed onto individual bodies;
those bodies are then sorted socially and in space;
the segregated bodies are then subject to
disparate opportunities and outcomes;
these unequal social groups then become the empirical basis
for racial categorization.
It is this vicious cycle that is the mechanism of systemic
inequality.
Rather than considering ``fairness'' to be a formal property
of a specific machine-learnt system, we propose that systems
can be designed with the objective of combating this cycle
directly and without reference to racial category.
Systems designed with the objective of integration of
different kinds of bodies
can discover segregated groups in an unsupervised way before using
fairness modifiers. 

Section \ref{sec:controversies}
outlines two controversies about race and machine
learning that have motivated research in this area.
We present these cases so that in subsequent sections
we can refer to them to illustrate our theoretical claims.

Section \ref{sec:political-categories} traces the history
of race in the United States from its roots in institutional
slavery and scientific racism through to changing demographic
patterns today.
This history reveals how race has always primarily been a
system for stigmatization which has only recently become the
site of ongoing political contest.
The racial categories continue to reproduce inequality.

Section \ref{sec:causes} discusses how racial categories
get ascribed onto individual bodies through identification
and classification. Seeing ascription as an event,
we identify several different causes for racial identity,
including phenotype, class, and ancestral origin.

Section \ref{sec:analysis} outlines the implications
of the history and sociology of race for system design.
We provide a heuristic for analyzing the software
and data of machine learning systems
for racial impact by categorizing them
as either colorblind, as explicit racial projects,
or as facilitators of users engaging in racial projects
(which may be racist, anti-racist, or neither).
We argue that designers have a dilemma:
using racial statistics reifies race, perpetuating
categories that are intrinsically unfair.
Not using them risks the systematic failures of
`color-blind' analysis, unwittingly reinforcing racial hegemony.

Section \ref{sec:new-design} offers a third alternative:
to design systems to be sensitive
to segregation in society across dimensions of phenotype,
class, and ancestral origin detected through unsupervised learning,
We sketch techniques for empirically identifying race-like
dimensions of segregation in both spatial distributions
and social networks.
These dimensions can then be used to group individuals for
fairness interventions in machine learning.

\section{Race and data: Controversies and context}
\label{sec:controversies}

In this section, we summarize two emblematic controversies
involving race and machine learning and some of the ensuing
scholarly debate.

\subsection{Recidivism prediction}

One area where racial bias and automated decision-making
has been widely studied is criminal sentencing.
The COMPAS recidivism prediction algorithm, developed by Northpointe,
was determined to have higher false positive rates for black defendants
than for white defendants \cite{larson2016we}, and charged with
being racially biased, even though explicitly racial information
was not used by the predictive algorithm \cite{angwin2016machine}.
This analysis has been contested on a variety of scientific
grounds \cite{flores2016false}, and the methodological controversy
has launched a more general interest in fairness in statistical
classification. Studies about the statistics of fair classification
have discovered that there is a necessary trade-off between classifier
accuracy and group based false positive and negative
rates under realistic distributions \cite{kleinberg2016inherent, corbett2017algorithmic}.
In light of the difficulties of interpreting and applying
anti-discrimination law to these cases \cite{barocas2016big},
a wide variety of statistical and algorithmic solutions to the
tension between predictive performance and fairness have been
proposed \cite{hardt2016equality, chouldechova2017fair}.
Reconsidering the problem as one of causal inference and the
predicted outcomes of intervention \cite{kilbertus2017avoiding, kusner2018causal},
especially in light of the purposes to which prediction and
intervention are intended \cite{barabas2017interventions} is a promising path forward.

The COMPAS algorithm did not explicitly use racial information
as an input.
It used other forms of personal information that were
correlated with race.
The fact that the results of an algorithm that did not
take race explicitly
into account were correlated with racial classifications
is an illustration
of the general fact that group-based disparate impact cannot be prevented
by ignoring the group memberships statistic; fairness must be accomplished
'through awareness' of the sensitive variable \cite{dwork2012fairness}.
Computer scientists have responded by
identifying methods for detecting the statistical proxies for a sensitive
attribute within a machine learnt model, and removing the effects
of those proxies from the results \cite{datta2017use}.

In this paper, we argue for a different understanding of the
role of racial categorization in the analysis of algorithms.
We argue that because of the socially constructed nature of race,
racial categories are not simple properties of individual persons,
but rather are complex results of social processes that are rarely
captured within the paradigm of machine learning.
For example, in the analysis of the algorithm that determined
alleged racial bias, the race of defendants was collected not
from the prediction software, but rather from the Broward County
Sheriff's Office: ``To determine race, we used the race classifications
used by the Broward County Sheriff's Office, which identifies defendants
as black, white, Hispanic, Asian and Native American.
In 343 cases, the race was marked as Other.'' \cite{larson2016we}
A focus on the potential biases of the
recidivism prediction algorithm has largely ignored the question of
how the Broward County Sheriff's Office developed its racial statistics
about defendants. We argue that rather than taking racial statistics
like these at face value, the process that generates them and the process
through which they are interpreted should be analyzed with the same rigor
and skepticism as the recidivism prediction algorithm. Thematically,
we argue that racial bias is far more likely to come from human judgments
in data generation and interpretation than from an algorithmic model,
and that this has broad implications for fairness in machine learning.

\subsection{Ethnic affinity detection}

Facebook introduced a feature to its advertising platform that allowed
the targeting of people in the United States based on racial distinctions,
which the company  called ``ethnic affinity'':
African American, Hispanic, or Asian American \cite{hern_2016}.
ProPublica discovered that this feature could be used to racially
discriminate when advertising for housing, which is illegal under the
Fair Housing Act of 1968 \cite{angwin2016facebook}. The report stated that
Facebook provided realtors with ad-targeting options that allowed
them to ``narrow'' their ads to exclude non-white groups like
Blacks, Asian, and Hispanics.  Facebook in fact drew the attention of the
Department of Urban Development (HUD) Secretary, Ben Carson who ordered an
investigation of Facebook's compliance with fair housing law.
Facebook decided
to pull the feature while also  increasing its certification of
advertiser's nondiscriminatory practices \cite{facebook2017improving}.

During the controversy, Facebook representatives explicitly made the
point that ``multicultural affinity'' was not the same thing as race.
It was not, for example, based on a user's self-identification with a race;
Facebook does not collect racial identity information directly.
Rather, ``multicultural affinity'' was based on data about users' activity,
such as the pages and posts they engaged with on the platform.

Indeed, the fact that racial groups could be profiled by
race despite not having users' individual racial identity data suggest
that race is much more than a characteristic of individual
identity, but rather is a socially reproduced form of categorical difference.
The feature did in fact give advertisers a tool to intentionally
or unintentionally engage in ``disparate racial treatment''.
However, pulling the feature did not make discrimination using
Facebook's platform impossible. \citet{speicher2018potential} have
investigated Facebook's advertising platforms and discovered that
even without the feature, there are ways to use the platform to
discriminate intentionally and
unintentionally, and propose that discrimination should be measured by its
effects, or its ``disparate impact''
\cite{feldman2015certifying, barocas2016big}.

Related work has been done on Google's advertising platform.
Discriminating ads have been delivered based on racialized search
results \cite{sweeney2013discrimination} and gendered user
profiles \cite{datta2015automated}. Studies about user
perception \cite{plane2017exploring} and legal
liability \cite{datta2018discrimination} have explored these issues in depth.
\citet{noble2018algorithms} argues
that digital media monopolies
like Google have engaged in ``algorithmic oppression''
that privilege white people and has led to both the
``commercial co-optation'' of black racial identities
as well as a kind of digital redlining against racial
minorities and women, especially Asian, Black,
and Latino women.
But the question remains: is this algorithmic oppression
simply the result of a white-dominated industry believing
that it was truly color blind, leading it to ultimately
ignore how race inequality might be reproduced digitally
and algorithmically, or are these instances of algorithmic
and digital racism systemic because searches and algorithms
mirror the racial beliefs of users?

\section{The political origin of racial categories}
\label{sec:political-categories}

Race is a social and cultural hierarchical system of categories
that stigmatizes differences in human bodies, but is not those
body differences themselves \cite{omi2014racial}. Race differences
are created by ascribing race classifications onto formerly
racially unspecified individuals and linking them to stereotyped
and stigmatized beliefs about non-white groups \cite{omi2014racial}.
We use \citet{link2001conceptualizing}'s definition of stigma as
``the co-occurrence of  labeling, stereotyping, separation (segregation), status debasement, and discrimination''.
For stigmatization to occur, power must be exercised \cite{link2001conceptualizing}.
Somebody is racially white not just because they have
less melanin in their skin, but because of the way society
has defined the societal rules for determining racial membership
and social status.

Folk conceptions of racial difference emerged in western
societies during the fifteenth century, but took on scientific
legitimacy during 17th and 18th centuries.
Lamarkian notions of natural and historic races gave way
to a more modern conception of race that emphasized the
immutability of the Blumenbach-inspired color-coded racial
groupings with which most of us have become familiar-
White, Black, Red, Yellow. Eighteenth century notions that
linked racial differences to environment and national
origin gave way to a more static conceptualizations
of racial difference now rooted biological deterministic
arguments (rooted in appearance) and Darwinism \cite{hattam2007shadow,gould1996mismeasure}.

Then ideologically supported by what is now debunked
scientific theory,
the white racial category has been nominally defined since
the first U.S. Census in 1790 named six demographic categories:
Free White males of 16 years and upward;
Free White males under 16 years;
Free White females;
All other free persons;
Slaves.
Known as ``The Naturalization Act'' on March 26, 1790, the
Senate and House of Representatives of the United States
of America enacted  ``An act to establish an uniform Rule of
Naturalization,'' and extended the possibility of citizenship
to ``any Alien being a free white person, who shall have
resided within the limits and under the jurisdiction of the
United States for the term of two years''
while naturalizing ``the children of citizens
of the United States that may be born beyond Sea, or out
of the limits of the United States.''
The Naturalization Act of 1790
insured that ``Free white persons'' would remain an officially
protected category for the next 160 years.

The construction of an Asian-American social category occurred
between the 1860's and 1960's. The State of Nevada was first
to pass anti-Asian legislation beginning in 1861, a precursor
to anti-miscegenation laws as as well congressional legislation
and judicial rulings that contributed to their social isolation
and social stigmatization \cite{sohoni2007unsuitable}.
In \emph{Takao Ozawa v. United States} in 1922, a
Japanese man who had
studied at the University of California and lived in the
United States was denied his request for citizenship
because he was ``clearly of a race which is not Caucasian.''
In \emph{United States v. Bhagat Sing Thind} (1923), a
``high-caste Hindu, of full Indian blood, born in Amritsar,
Punjab, India'' was denied citizenship because, though
Caucasian, he was not white.
The Immigration and Nationality Act of 1952 (known as
The McCarran-Walter Act) removed racial restrictions in
Asian naturalization, while it also created an Asian quotas
system based on race rather than on nationality \cite{tsuda2014m}.

In the case of Negro (Black) Americans, those once
categorized by the US Census as either ``Slave'' or
``Other free persons'', their racial classification varied.
The 1870 and 1880 Censuses recognized mixed-race persons as ``mulatto''
(defined as someone who is Negro and at least one other race),
while the 1890 Census added another mixed category,
``quadroon,'' to refer to persons who were ``one-fourth black blood.''
\emph{Plessy v. Ferguson} (1896) established the legality of
racial segregation that was `separate but equal', as well
as confirmed the quantification of race by law.
And while the 1900 Census dropped all but the Negro category,
the 1910 and 1920 Census briefly brought back the mulatto category
only to drop it one final time in the 1930 Census, which finally
solidified the Negro category once and for all along the lines
of the ``one drop rule'', meaning that a single drop of
``African blood''
was sufficient to make a person ``Negro.''
This rule was called the hypodescent rule by anthropologists
and the ``traceable amount rule'' by the US courts.
``It was policed by an array of government agencies,
market practices, and social norms and was
ultimately internalized by individuals of mixed
European and African lineage'' \cite{haynes2018soul}
In fact, the one drop rule treated blackness as a
contaminant of whiteness, thus granting rights to those
deemed white and, by definition, privileged.

After World War II, the political trajectory of race in
the United States evolved.
President Truman abolished discrimination based on race,
color, religion, or national origin in the armed forces with
Executive Order 9981.
Scientific racism fell into disfavor among scientists
and scholars after the war as it was strongly associated with
the defeated German Nazis.
A new social theory of ethnicity, that attempted to reduce
racial difference to cultural difference and emphasized the
possibility of assimilation and equality, became increasingly
popular.
But while this accounted for the assimilation of many new
immigrant groups, real segregation and stratification along
race lines ensured that racial categories remained ingrained
in societal consciousness, including the anti-racist consciousness
that mobilized for equal rights.
Racial categories that had once been a political myth had solidified
into social fact through the mechanisms of segregation.

The Civil Rights Movement in the 50's and 60's lead to
\emph{Brown v. Board of Education}, which ended \emph{de jure}
racial segregation, and
the passing of anti-discrimination laws such as the
Voting Rights Act of 1965 and the Civil Rights Act of 1968,
which prohibited discrimination based on race.
Thus anti-discrimination law reified the same racial categories
that had been defined as a tool for subjugation and segregation.
Census data would then track racial statistics partly in order
to enforce civil rights laws.
Anti-discrimination policies have in the years since they
have passed provoked ``racial reaction'' as whites have
rearticulated their political interests in new ways.
Racist and anti-racist political currents have been dialectically
battling over racial policy since the rise of anti-racist
consciousness.

\citet{omi2014racial} characterize the changes to racial
categories through political contest as ``racial formation''.
Key to their theory of racial formation is the ``racial project'',
``The co-constitutive ways that racial meanings are translated into
social structures and become racially signified.''

\begin{definition}[Racial project]
  ``A racial project is simultaneously an interpretation,
  representation, or explanation of racial identities and meanings,
  and an effort to organize and distribute resources
  (economic, political, cultural) along particular racial lines.'' \cite{omi2014racial}
\end{definition}

Racial projects can be racist,
anti-racist, or neither depending on how they
align with ``structures of domination based on racial significance and identities''.
Racial categories in the 21st century are the result of an
ongoing contest of racial projects that connect how
``social structures are racially signified'' and
``the ways that racial meanings are embedded in social structures'',
thereby steering state policy and social practice.

These policy changes have had lasting changes on the
demographics and spatial and social distribution of
the population of the United States.
This has allowed racial categories to change, albeit slowly,
as each generation experiences race differently. For example,
the United States Supreme Court struck down laws banning
interracial marriage in 1967 with \emph{Loving v. Virginia}.
When interracial parents desired for their children
mixed-race identity, they put political pressure on
institutions to recognize their children as such.
In 2000, the option to mark
``one or more'' racial categories was adopted by the Census
in 2000.

\section{Causes of racial identification and ascription}
\label{sec:causes}

Racial categories fill societal imagination and are
solidified by law. But racial categories have their
effect by being ascribed to individual bodies.
Because it is not an intrinsic property of persons but
a political category, the acquisition of a race by a person
depends on several different factors, including biometric
properties, socioeconomic class, and ancestral geographic
and national origin.

\subsection{Biometric properties}

Though the relationship between phenotype and race
is not straightforward, \citet{omi2014racial} argue that there
is an irreducible ocular-corporeal component to race:
race is the assigning of social meanings to these visible
features of the body. Indeed, the connection between phenotype
and race has been assumed in research on fairness in machine
learning. In their work on intersectional accuracy disparities
in gender classification based on photographs of faces,
\citet{buolamwini2018gender} use the dermatologist approved
Fitzpatrick Skin Type classification system to identify faces
with lighter and darker skin. While they draw the connection
between phenotype and race, they note that racial categories
are unstable and that phenotype can vary widely within a racial
or ethnic category. Indeed, it is neither the case that race can
be reduced to phenotype, nor that phenotype can be reduced to
race: there is broad empirical evidence that shows that intraracially
among people identified as black, the lighter skinned are treated
favorably by schools and the criminal justice system compared
to those with darker skin \cite{burch2015skin, viglione2011impact, monk2014skin}.

Phenotype is a complex consequence of genotype, which is in turn a consequence of biological ancestry. With commercially available genetic testing, genotype data is far more available than it has been historically. It has also exposed the fact that many people have ancestry that is much more ``mixed'', in terms of politically constructed racial categories, than they would have otherwise assumed; this has had irregular consequences for people's racial identification \cite{roth2018genetic}.

Both phenotype and genotype may be considered biometric properties under the law, and hence these data categories would be protected in many jurisdictions. However, despite these protections, this data is perhaps more available than ever. Personal photographs, which can reveal phenotype, are widely used in public or privately collected digital user profiles.

\subsection{Socioeconomic class}

While racial categories have always been tied to
social status and economic class, the connection
and causal relationship between race and class
has been controversial. \citet{wilson1978declining} argued
that in the postwar period the rise of the black elite
and middle class made race an issue of ``declining significance'',
despite the continued existence of a black underclass.
\citet{omi2014racial} are critical of this view,
noting the fragility of the black middle class and
its connection to the vicissitudes of available public
sector jobs.
Recent work by \citet{chetty2018race} on racial effects
of intergenerational class mobility confirm that black
children have lower rates of upward mobility and higher
rates of downward mobility compared to white children,
even when controlling for those that ``grow up in two-parent
families with comparable incomes, education, and wealth;
live on the same city block; and attend the same school.''
This is due entirely to differences in outcomes for men,
not women. \citet{massey2007categorically} accounts for
the continued stratification along racial lines as a
result of ingrained, intrinsic patterns or prejudice,
which is consistent with \citet{bordalo2016stereotypes}.
In the controversial work of \citet{saperstein2012racial},
racial self-identification and classification was found to be
fluid over
time in reaction to changes in social position, as signaled
by concrete events like receiving welfare or being incarcerated.
This effect has been challenged as a misinterpretation of
measurement error \cite{kramer2016racial}, though similar results have surfaced outside of the U.S., as when Hungarians
of mixed descent are more likely to identify as Roma if under
economic hardship \cite{simonovits2016economic}.

Confounding the relationship between individual race and class
is the fact that socioeconomic class is largely inherited;
in other words, there is always some class immobility.
This is acknowledged both in economics in discussions of
inherited wealth (e.g. \cite{piketty2014capital}) and
more broadly in sociology with the transfer of social
capital via the family and institutions that bring similar
people together with the function of exchange \cite{bourdieu2011forms}.
The ways that racial social and spatial segregation lead to the
``monopolistic group closure'' of social advantage on racial lines is discussed in \citet{haynes2008place}.

\begin{figure}
\begin{center}
\begin{tikzcd}
  Genealogy \arrow[d] \arrow[dr] \arrow[drr] & & \\
  Genotype \arrow[d] & Inheritance \arrow[d] & Nationality \arrow[d] \\
  Phenotype \arrow[drr] & Class \arrow[dr] & Categories \arrow[d] \\
  &  & Race \\
\end{tikzcd}
\end{center}
\caption{A model of how individual biological properties
  (genealogy, genotype, and phenotype) are
  racialized through national political categories
  and associations with socioeconomic class.
  Here inheritance refers to all forms of capital, including
  economic and social, passed from parents to children.
  Broadly speaking, genealogy is a strong determiner of race,
  but importantly as a common cause
  of phenotype, class, and nationally recognized racial categories,
  which are separate
  components of racial classification.
}
\end{figure}
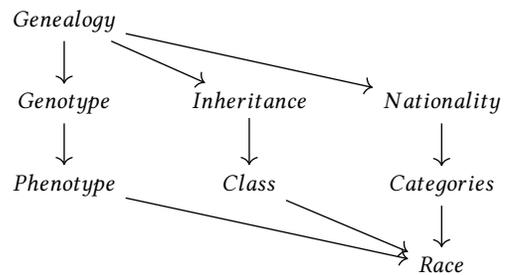

\subsection{Ancestral national and geographic origins}

The definitions of races used in the U.S. census are
``rife with inconsistencies and lack
parallel construction'' \cite{omi2014racial}, but have
nevertheless become a de facto standard of racial and
ethnic classification.
Blacks are defined as those with ``total or partial
ancestry from any of the black racial groups of Africa''.
Asian Americans are those ``which have ancestral origins
in East Asia, Southeast Asia, or South Asia''.
Hispanic Americans are ``descendants of people from
countries of Latin America and the Iberian Peninsula,''
and is considered by the census as an ethnicity and not
a race.
Though phenotype and class may be social markers of race,
beliefs about race as a ``true'' intrinsic property are
anchored in perceptions and facts about personal ancestry.

Ancestry
is one of the main conduits of citizenship, which
determines which legal jurisdiction one is subject to.
These jurisdictions can influence what categories a person
individually identifies with.
It is not only in the United States that racial categories
are anchored in ancestry, even though racial categories
are constructed differently elsewhere.
\citet{loveman2014national} notes that Latin American
nations with a history
of slavery commonly use the Black racial category,
whereas those with without that history are socially
more organized around Indigeneity.
Racial categorization anywhere will depend on those
categories available by legal jurisdiction;
this can be striking to those who migrate and find
themselves ascribed to something unfamiliar.
(Consider a person of Latin American of European ancestry who,
upon moving to the United States, becomes a Hispanic.)

\section{Heuristics for analysis and design of systems}
\label{sec:analysis}

We now address how the history and social theory of race discussed
above applies to the design of machine learning and
other computer systems.
Responding to the provocation raised by
\citet{noble2018algorithms}, we argue that there is a substantive
difference between systems that result in a controversial
or unfair outcomes due to the racial bias of their
designers and those that do so because
they are reflecting a society that is
organized by racial categories.
Using the concept of a ``racial project'' introduced
in Section \ref{sec:political-categories},
we propose a heuristic for detecting racism in machine
learning systems.

We draw a distinction between the software used by a
machine system and its input and output data.
Further, we distinguish between three categories of systems
that are not mutually exclusive: those that are themselves
racial projects, those that allow their users to engage
in racial projects, and those that attempt to be ``blind''
to race.
Racial projects may be racist, anti-racist, or neither.

Machines that attempt to correct unfairness
through explicit use of racial classification do so at the risk
of reifying racial categories that are inherently unfair.
Machine learning systems that allocate resources in ways
that are blind to race will reproduce racial inequality in
society.
We propose a new design in Section \ref{sec:new-design}
that avoids both these pitfalls.

\begin{figure}
\begin{center}
  \begin{enumerate}
  \item How has the software designed?
    \begin{itemize}
    \item ``Blind'' to race. (A)
    \item As a racial project. (B)
    \item Enabling users' racial projects. (C)
    \end{itemize}
  \item Are the input data racialized?
    \begin{itemize}
    \item Not explicitly. (A)

    \item Explicitly, by ascription. (B)
    \item Explicitly, by self-identification. (C)
    \end{itemize}
  \item Is the system output racialized?
    \begin{itemize}
    \item Not at all. (A)
    \item System ascribes race. (B)
    \item By user interpretation. (C)

    \end{itemize}
\end{enumerate}
\end{center}
\caption{Heuristics for analysis and design of algorithmic systems.
  Systems of type A are ``blind'' to race and therefore risk
  learning and reproducing the racial inequality inherent in society.
  Systems of type B explicitly use ascribed racial labels,
  and so risk reifying
  racial categories by treating race as an intrinsic property
  of a person.
  These systems are racial projects, in the sense that
  they represent racial
  categories in a way that is relevant to resource allocation.
  Systems of type C may be considered racial projects,
  but have the distinction that they enable the system users
  to engage in their own racial projects.
  Racial projects (whether in type B or type C systems)
  may be racist, anti-racist, or neither, depending
  on how they align with structures of domination in society.
  Categories A, B, and C are not mutually exclusive;
  they are distinguished here as analytic heuristics only.
}
\end{figure}

\subsection{How has the software been designed?}

A first step to evaluating the racial status of a machine system
is to evaluate whether the software it uses has been designed for
the purpose of achieving a racial outcome or representation.
Using the language of  \citet{omi2014racial}, the question is
whether or not the software has been designed as a racial project.

If software has been designed as a racial project,
then it is appropriate to ask whether or not the
racial project is racist, anti-racist, or neither.
A racial project is racist, according to \citet{omi2014racial},
``if it creates or reproduces structures of domination based on
racial significance and identities,'' and anti-racist if it
``undo[es] or resist[s] structures of domination based on
racial significations and identities.''

\begin{example}
In the case of Facebook's ethnic affiliation feature, Facebook engaged in a racial project: to discover and represent the racial affiliations of its users. Doing so was neither a racist nor an anti-racist project. That it passed these representations on to the users of its advertising platform gave advertisers the ability to engage in broad range of racial projects. These possible racial projects included the potential for illegal racist discrimination in housing advertising.
\end{example}

The criterion for system software engaging in a racial project
is that it engages
racial categories through the words, concepts, or
social structures that
abstractly represent racial differences.
Racial projects are efforts to change these categories
in one way or another.
Many systems that use machine learning are also,
by design, platforms for their users' political expression.
These platforms perhaps inevitably become fora for
their users' diversely racist, anti-racist, and other
racial projects.

We have also seen that not all institutional outcomes with disparate
racial impact are due to racist racial projects;
even `colorblind' institutions can have disparate outcomes for groups
of people that identify with or are ascribed race based on
racial categories. Software that has not been designed with any
intentional reference to race may still treat people who identify
as black relatively poorly.
These systems, which correspond roughly with institutions of racial
hegemony critiqued by \citet{omi2014racial}, reflect a status quo
of racial inequality without engaging in it. 

\subsection{Are the input or output data racialized?}

Beyond the mechanics of the system's software, we can also
evaluate a system's input data and output.
Is the input or output being racialized? If so how?

Input data to a machine learning system, especially if it's
personal information, may have explicit racial labels.
These may be generated from individual self-identification,
institutional ascription, or both. As discussed above,
both self-identification and institutional classification
are socially embedded and changeable based on circumstance.
These labels by definition place individuals within a political
schema of racial categorization. As such, it is a mistake to
consider such labels a ``ground truth'' about the quality of
a person, as opposed to a particular event at a time, place, and context.

Every instance of racial classification in input data should therefore,
as a matter of sound machine learning practice, be annotated with
information about who made the ascription, when, and under what circumstance.
To do otherwise risks reifying race, treating a person's ascribed race
as an intrinsic feature, which unfairly places them within a system of
inequality \cite{zuberi2000deracializing}.
It does this even if the ultimate use of the data is an
anti-racist racial project; indeed, the potential for racist use of this data is always available as an exposure threat.

In addition, this annotation may give analysts clues as to the political
motivations of the system designers and data providers.
Political context should be seen as part of the generative process
that must be modeled to best understand data sources.
For example, the degree to which somebody has culturally assimilated,
or the degree to which a ``one drop rule'' of racial classification
or recognition of multi-racial identity is in effect, may be an
important factor in determining the distribution of racial labels.

We propose that as a heuristic for analyzing a system for its
racial impact, an analyst attend to whether the inputs and outputs
of the system are racialized either (a) explicitly through the
\emph{ascription} of racial categories, (b) explicitly through either
the self-identification or subjective \emph{interpretation} of the user,
or (c) not at all.
Those systems designed for ascription are likely to be themselves
racial projects, in that they use racial categories by design.
Systems whose inputs allow for racial self-identification may also
be racial projects, but also crucially allow for their users to
engage in racial projects using the system based on how they
represent themselves as a member of a race.

System's whose outputs are racialized by user interpretation
may not be racial projects themselves; however, users can engage
in racial projects based on how the system represents other people.
Because race is an ascribed category, users of a system
can ascribe race to people
represented by a system based on ocular cues, dress,
and other contextual information.
Especially if these representations accord with racial
stereotypes \cite{bordalo2016stereotypes},
there may be the perception that the system is reproducing
racial disparities.
If the outputs of a system are racialized by interpretation
but not explicitly, that interpretive discourse can itself be a racial project.
In other words, the outputs of a system, such as a search engine, can be the subject
of a conversation about race and resource allocation more generally.
However, it may be an error to attribute the content of a
racialized interpretation of a system to the system itself.
A thorough analysis of the system inputs, software, and outputs is necessary to determine
where racial intent or social racial categories caused the output or ascription.

Some systems whose input and output data represent people may not be explicitly
racialized at all.
However, since racial categories structure inequality
pervasively throughout society, these systems will likely reproduce
racial inequality anyway.
The difficulty of designing a system that neither reproduces racial social
inequality nor reifies racial categories, which are inherently unfair,
motivates an alternative design discussed in the next section.

\section{An alternative: designing for social integration}
\label{sec:new-design}

System designers of machine learning systems that determine resource
allocation to people face a dilemma. They can ignore racial inequality in society, and risk having
the system learn from and reproduce systemic social inequality due to racial
categorization. They can also use racial statistics to try to mitigate
unfairness in outcomes, but in doing so they will reify racial categorization.
We present a third option as a potential solution to this dilemma.

\begin{figure}
\begin{center}
\begin{tikzcd}
 & Ascription \arrow[dr] & \\
 Formation \arrow[ur] & & Sorting \arrow[dl] \\
 & Disparity \arrow[ul] & \\
\end{tikzcd}
\end{center}
\caption{Schematic of vicious cycle of racial formation.
  Bodies are ascribed into racial categories,
  then sorted socially and in space based on those ascriptions.
  These sorted bodies are then exposed to disparate outcomes.
  Racial categories are then formed on the basis of those unequal outcomes
  and their distribution across people based on phenotype,
  ancestry, and class indicators.
  Those categories are then ascribed to bodies, repeating the cycle.
}
\label{fig:cycle}
\end{figure}
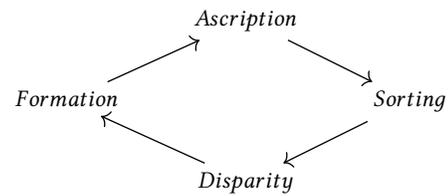

This proposal rests on two theoretical assumptions.
First, recall that in our outline of the formation of race, segregation and
stratification of populations play a key systemic role.
The social fact of racial categorization is reinforced through
policies and practices of segregation and stratification.
Racial categories are ascribed onto individual bodies;
those bodies are then sorted socially and in space;
the segregated bodies are then subject to disparate opportunities and outcomes;
these unequal social groups then become the empirical basis for racial categorization (see Figure \ref{fig:cycle}).
Rather than consider ``fairness'' to be a formal property of
a specific machine-learnt system,
we propose that systems can be designed to disrupt this vicious cycle.
This requires treating groups that have been segregated socially and
in space similarly, so that disparate impacts do not apply.

The second assumption addresses the problem that ascribing
politically constructed racial categories reifies them,
which contributes to status inequalities.
We ask: how can systems designed with the objective of integration
of different kinds of bodies, especially those bodies that have been
sorted racially, but without reference to racial categories themselves?
Our alternative design also draws on our conclusion from
Section \ref{sec:causes}.
The facts about people that cause ascription and self-identification
with politically constructed status categories are facts
about phenotype, social class (including events that signal social position),
and ancestry.
We propose that categories reflecting past racial segregation
can be inferred through unsupervised machine learning based on these
facts.
These inferred categories can then be used in fairness modifiers
for other learning algorithms.

By using inferred, race-\emph{like} categories that are adaptive to
real patterns of demographic segregation, this proposal aims
to address historic racial segregation without reproducing
the political construction of racial categories.
A system designed in this way learns based on real demographics
of the populations for which they are used,
and so will not result in applying national
categories in a context where they are inappropriate.
It is also adaptive to demographic changes in the same place or social
network over time.

\subsection{Detecting spatial segregation}
\label{sec:detect-spatial}

Spatial segregation into different neighborhoods
is one of the main vehicles of disparate impact
on people of different races.
In part because of the its long history,
the question of how to best measure spatial segregation
is its own subfield of quantitative sociology
\cite{white1983measurement,reardon2004measures,wong2005formulating}
whose full breadth is beyond the scope of this paper.
The most basic measure, which is both widely used and
widely criticized, is the dissimilarity measure, $D$.

\begin{definition}[Dissimilarity (Black and white)]
  $$D = \frac{1}{2}\sum_i \left \vert \frac{w_i}{W} - \frac{b_i}{B}  \right \vert$$
  where $i$ ranges over the index of spatial tracts, $w_i$ and $b_i$
  are the white and black populations in those tracts,
  and $W$ and $B$ are the total white and black populations
  in all tracts.
\end{definition}

While defined above with respect to only two racial groups,
generalized versions of the metric have been proposed for multiple
groups.
Most criticism of this metric is directed at the fact that it is
``aspatial'', obscuring true spatial relationship through the division
of land into parcels, which may be done in a way that invalidates the
result \cite{white1983measurement,reardon2004measures,wong2005formulating}.
For the purposes of this article, we will assume that the spatial
tracts are selected adequately in order to focus on a different
criticism: that this metric assumes that the population has
been ascribed to racial categories, thereby reifying them.
We propose a modification of this metric for identifying
race-\emph{like} categories of segregation between land tracts.

Consider the following sketch of method of detecting spatial segregation.
Let $i$ range over the indices of land tracts.
Let $j$ range over the indices of individuals.
Let $\vec{x}_j$ be a vector of available personal
data about each individual,
including information relevant to phenotype (perhaps
derived from photographs), class, and national origin.
For simplicity, consider the vector to be of binary features.
Let $i(j)$ be the index of the tract where person $j$ resides.
Let $\vec{X}_i$ be the aggregation (by summation) of all $x_j$
such that $i(j) = i$ as a normalized vector.
Let $\mathbf{X}$ be all $\vec{X}_i$ combined into a matrix.

The first principle component of $\mathbf{X}$ will be a feature
vector in the same space as the parcel data vectors $\vec{X}_i$
that reflects the dimension of greatest variance between parcels.
Because the parcel data vectors aggregate information about the
components of race (phenotype, class, and nationality), this would
reflect racial segregation without depending on any particular historical
or political racial categorization.
Other principle components would likewise reflect other elements of
racial segregation.
Persons could then be racially classified by transforming their
personal data vector through the principle component and thresholding
the result.
This classification could then be used as an input $A$ to fairness
interventions in machine learning.

\subsection{Detecting social segregation}

Social segregation by race may be operationalized using
network representations of society.
Homophily, the phenomenon that similar people are more likely
to be socially connected, is a robustly studied and confirmed result
\cite{mcpherson2001birds}, and the problem of bridging between
isolated niches has been posed as a general social problem beyond
the context of race \cite{diprete2011segregation}.
Several metrics for measuring social segregation of all kinds
have been proposed
\cite{freeman1978segregation}
\cite{bojanowski2014measuring}.
These metrics have in common that they assume that nodes in the network
have already been accurately assigned to different groups.

One widely known segregation measure for discrete properties is
the assortativity coefficient \cite{newman2003mixing}, defined as:

\begin{definition}[Assortativity coefficient]
  $$r = \frac{\sum_i e_{ii} - \sum_i a_i b_i }{1 - \sum_i a_i b_i}$$

  where $e_{ij}$ is the fraction of edges in the network that connect
  nodes of group $i$ to nodes of group $j$, $a_i = \sum_j e_{ij}$,
  and $b_j = \sum_i e_{ij}$.
\end{definition}

As $r$ approaches $1$, the network gets more assortatively mixed,
meaning that edges are within group.
If the groups in question are racial classifications, an
assortatively mixed network is a segregated network.

To adapt to the case where racial classification is not
given, but component racial features such as phenotype,
class, and nationality are available as vector $\vec{x}_i$
(again, of binary features, for simplicity)
consider a method similar
to that proposed in Section \ref{sec:detect-spatial}.
For each edge between $i$ and $j$, aggregate $\vec{x}_i$
and $\vec{x}_j$ into $\vec{X}_{ij}$ by summing them, then
combine these into
a matrix $\mathbf{X}$, and use the principle components
to determine the dimensions of greatest variation between
the aggregated properties of each connected pair.
As before, transforming the individual feature
vectors by the components and applying a threshold
then assigns each person to the race-like groups
of greatest social segregation.
Measuring the assortativity coefficient for these
groups will provide another measure of the
segregation of the population along race-like lines.
These classifications can then be used as protected
groups $A$ in fair machine learning.

\subsection{Threats to validity and future work}

We have proposed that as a normative goal,
systems can be designed to promote similar treatment
of bodies that are otherwise segregated socially
or in space.
This proposal is motivated by social theory of
how segregated perpetuates racial categories
as a system of status difference.
We have not implemented or tested this design
and here consider threats to its validity.

An empirical threat to its validity is if
the principal components of the aggregated feature matrices
do not reflect what are recognizable as racial categories.
This could be tested straightforwardly by collecting
both ascribed (or self-identified) racial labels and
other features for a population and computing how well
the principle component vectors capture the ascribed
racial differences.
If the categories were not matched, then it could be
argued that the system does not address racial inequality.

On the other hand, if there are ever race-like dimensions of
segregation that have not been politically recognized as
racial categories, then that is an interesting empirical
result in its own right.
It suggests, at the very least, that there are active forms
of discrimination in society based on properties of
people that are not currently recognized politically.
We see the discovery of potentially unrecognized forms
of discrimination as a benefit of this technique.

Another threat to validity of our analysis is the known
fact that the schematic ``vicious cycle'' of racial
formation presented in Figure \ref{fig:cycle} is
an over-simplification.
We have drawn this theory from a survey of sociology
literature on the formation of race.
However, we now only have a hypothesis about the actual
effects of such a system designed
as proposed here on the politics of race over time.
Confirming that hypothesis will require implementation
and extensive user testing, perhaps through a longitudinal study.

Our discussion of strategies and metrics for reducing
segregation along race-like lines has been brief
due to the scope of this paper.
We see refinement of these techniques as a task
for future work.
An example open problem raised by the preceding discussion
is which social network segregation measures are best
at capturing the effects of racial inequality.

\section{Discussion}

Controversies surrounding machine learning's treatment
of race have inspired a growing field of research about
fairness in machine learning.
This field often treats fairness as formal property of
computational systems, where fairness is evaluated in terms
of a set of protected groups ($A$, in our notation).
The system is considered fair if outcomes are in some sense
balanced with respect to the groups.
Group membership is considered a simple fact about natural persons.

In this paper, we have scrutinized what it means for racial identity
to be a 'protected group' in machine learning.
We trace the history of racial categories in U.S. law and
policy to show how racial categories became ingrained in society
through policies of segregation and exclusion.
The recent manifestation of them in civil rights law is still
based on their role as political status categories ascribed based on
differences in body, class, nationality, and ancestral origin.
Because they are intrinsically categories of disadvantage and
inequality, there is nothing fair about racial identity.

System designers are caught in a dilemma: ignore race and reproduce
the inequality of race by accident, or explicitly consider racial
statistics in order to mitigate inequality in favor of fairness.
Through its use of racial classification, the latter systems
put themselves in a paradoxical position of making the unequal
equal, and invite political opposition and cooption.

We propose a third way based on the insight that racial categories
are perpetuated by real patterns of segregation in space and society.
We argue that rather than promote ``fairness'' as a system property,
systems should be designed with the objective of promoting social integration based on similar treatment of segregated populations.
To perform this function, systems need a way to determine which which
populations are racially segregated without reifying existing
racial categories by dependence on racial statistics.
We propose unsupervised learning methods for finding latent
dimensions of racial segregation in race and society.
These dimensions can be used to dynamically classify people into
situationally sensitive racial categories that can then be
entered into fairness computations.

Racial categories, and the disadvantage associated with them, are solidified through segregation in housing, education, employment, and civic life, which can happen through legislation and institutional mechanisms.
It is the segregation and the disparate advantages of being in segregated groups that is the cause of unfairness.
We are proposing that ``fairness in machine learning'' should be designed to detect segregation in an unsupervised way that does not reify the historical categories of reification while nevertheless being sensitive to the ongoing effects of those categories.
This design is adaptive to social change, the emergence of new segregated and discriminated-against groups, and also the emergence of new norms of equality.

\bibliographystyle{ACM-Reference-Format}
\bibliography{racial-information.bib}

\end{document}